\documentclass[10pt,twocolumn,letterpaper]{article}

\usepackage[pagenumbers]{cvpr} 

\usepackage{graphicx}
\usepackage{amsmath}
\usepackage{amssymb}
\usepackage{booktabs}

\usepackage{multirow}
\usepackage{tabularx,verbatim}
\usepackage{xspace}
\usepackage{algorithm}
\usepackage{algorithmic}
\usepackage{setspace}
\usepackage{comment}
\usepackage{bbm}
\usepackage{enumitem}
\usepackage{colortbl}
\usepackage{color, xcolor} 
\usepackage[T1]{fontenc}
\usepackage{bbm}
\usepackage{listings}
\usepackage{subcaption}
\usepackage{pifont}
\usepackage{multicol}
\usepackage{soul}
\usepackage{fontawesome}
\usepackage[symbol]{footmisc}

\definecolor{remark}{rgb}{1,.5,0} 
\definecolor{citecolor}{rgb}{0,0.443,0.737} 
\definecolor{linkcolor}{rgb}{0.956,0.298,0.235} 
\definecolor{cyan}{rgb}{0.831,0.901,0.945}

\usepackage{amsmath}

\definecolor{cvprblue}{rgb}{0.21,0.49,0.74}
\definecolor{mycolor}{HTML}{EAB48A}
\definecolor{black}{RGB}{219, 48, 122}
\definecolor{magenta}{rgb}{0.73, 0.31, 0.56}
\definecolor{mustard}{rgb}{0.99, 0.65, 0.0}
\definecolor{myred}{RGB}{186,79,143}
\usepackage[pagebackref,breaklinks,colorlinks,citecolor=cvprblue,bookmarks=false]{hyperref}

\def\paperID{3981} 
\def\confName{CVPR}
\def\confYear{2024}

\newcommand{\ours}{Animate124\xspace}

\title{\ours: Animating One Image to 4D Dynamic Scene}

\author{%
Yuyang Zhao$^1$,
Zhiwen Yan$^1$,
Enze Xie$^2$\textsuperscript{\faEnvelopeO}, 
Lanqing Hong$^1$,
Zhenguo Li$^3$,
Gim Hee Lee$^1$ \\
{$^1$ National University of Singapore}  \\
{$^2$ The University of Hong Kong }\\ 
{$^3$ The Hong Kong University of Science and Technology}\\ 
\textbf{\url{https://animate124.github.io/}}
\vspace{-.1in}
}

\begin{document}

\twocolumn[{
      \maketitle
      \begin{center}
        \centering
        \vspace{-0.1in}
        \includegraphics[width=.99\linewidth]{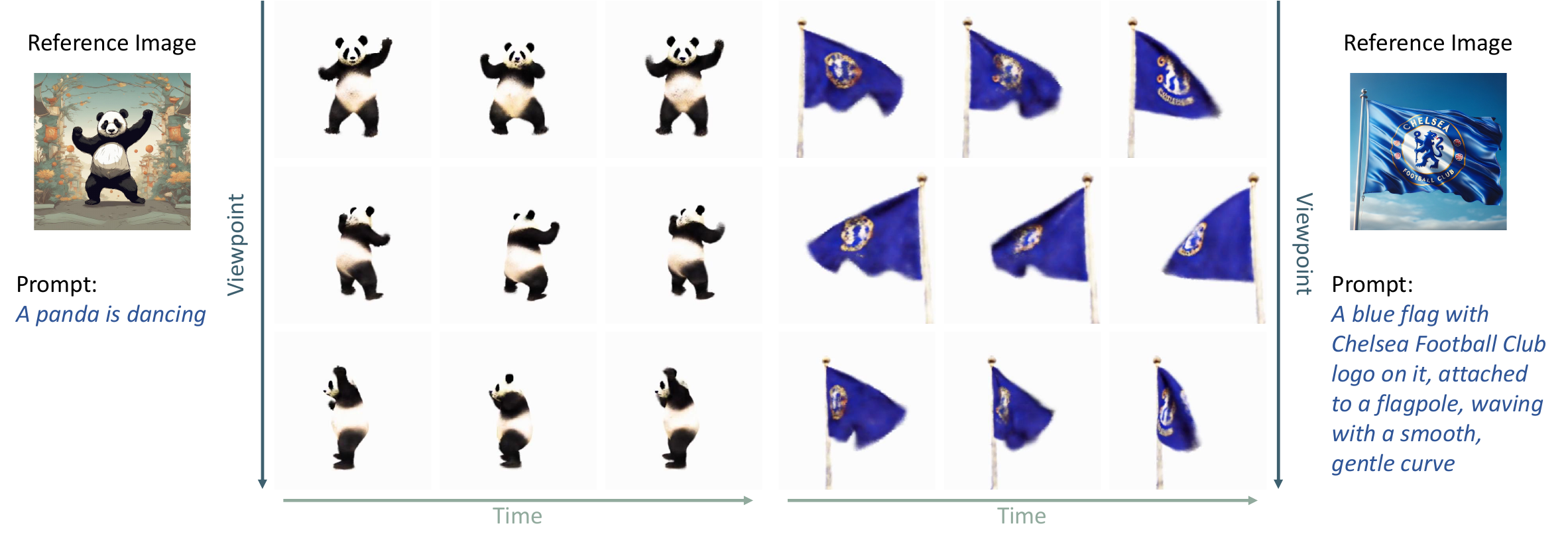}
        \vspace{-.1in}
        \captionof{figure}{{To the best of our knowledge, our}
          \textbf{\ours} is the first framework to animate one in-the-wild image into 3D video with the motion defined by text prompt.
        }
        \label{fig:teaser}
      \end{center}
    }]

\begin{abstract}
\footnotetext{\faEnvelopeO~ Corresponding author.}

    We introduce \ours (\textbf{Animate}-\textbf{one}-image-\textbf{to}-\textbf{4}D), the first work to animate a single in-the-wild image into 3D video through textual motion descriptions, an underexplored problem with significant applications. Our 4D generation leverages an advanced 4D grid dynamic Neural Radiance Field (NeRF) model, optimized in three distinct stages using multiple diffusion priors. Initially, a static model is optimized using the reference image, guided by 2D and 3D diffusion priors, which serves as the initialization for the dynamic NeRF. Subsequently, a video diffusion model is employed to learn the motion specific to the subject. However, the object in the 3D videos tends to drift away from the reference image over time. This drift is mainly due to the misalignment between the text prompt and the reference image in the video diffusion model. In the final stage, a personalized diffusion prior is therefore utilized to address the semantic drift. As the pioneering image-text-to-4D generation framework, our method demonstrates significant advancements over existing baselines, evidenced by comprehensive quantitative and qualitative assessments.
\end{abstract}

\section{Introduction}

Generative models have recently achieved significant advancements in producing 2D images~\cite{LDM,Imagen,DALLE-2} and videos~\cite{makeavideo,luo2023videofusion,zhou2022magicvideo}. The creation of 3D content~\cite{DreamFusion,DreamBooth3D,magic123}, a critical representation of the real world, has garnered increasing attention and has shown rapid development. Empowered by robust priors from both 2D and 3D diffusion models, the generation of 3D scenes~\cite{Text2Room,DreamFusion,RealFusion,Zero-1-to-3,mav3d} from text prompts or single images is now feasible and has witnessed remarkable progress. However, most existing research concentrates on static scenes %
{and often overlooks} the dynamic nature of the real world.

In contrast, dynamic 3D scenes (or 3D videos) more effectively represent the richly informative 3D world, offering significant applications in video games, augmented reality, and virtual reality. %
{Despite its importance, 3D video generation remains relatively unexplored partly due to the under-development in video generation models.}
MAV3D~\cite{mav3d} represents a pioneering effort in generating 3D videos from text prompts. It employs temporal Score Distillation Sampling (SDS)~\cite{DreamFusion} to transfer knowledge from a text-to-video diffusion model~\cite{makeavideo} into a dynamic NeRF representation~\cite{hexplane}.
However, content generation solely from text 
{often lacks control despite its diversity}. This limitation has inspired the integration of various conditioning signals in 2D~\cite{controlnet,huang2023composer,wang2023videocomposer} and image-based conditioning in 3D~\cite{magic123,RealFusion,NeuralLift}.
In the realm of controllable video generation, the technique of conditioning on an initial image to guide subsequent motion~\cite{wang2023videocomposer,gen1} has become both effective and popular.
For {example}, generating a scene based on the prompt: ``A panda is dancing'', users may wish to define the appearance and starting pose {of the panda} by providing a reference image.

In this paper, 
{we explore image-text-based 3D video generation having inspired by text-to-3D-video and controllable video generation.}
{Guided by the motion outlined in a text prompt, our approach aims to lift and animate a single input image into a 3D video.}
To achieve this {goal} with limited input, we utilize {Score Distillation Sampling} (SDS)~\cite{DreamFusion} to infuse knowledge from diffusion priors into a dynamic NeRF model.
{This implies that} an efficient and effective NeRF model is essential for representation. MAV3D employs HexPlane~\cite{hexplane} to map the X, Y, Z, and time axes onto six 2D planes 
{and} subsequently fusing these features to determine density and color. 
{Nonetheless, we risk falling into the Janus problem with a naive adoption of MAV3D since we only have a single reference image as both a condition and a source of supervision. Specifically, the Janus problem happens when the camera direction of the reference image view is perpendicular to one of the 2D planes in HexPlane. 
As a consequence, the plane features overfit to the reference image, where both the front and back views resemble the reference image without a proper 3D geometry.
}
{Moreover, we empirically observe some traces of the Janus problem even when the reference image is off-perpendicular the 2D planes in HexPlane.}
{Refer to Appendix.~\ref{sec:hexplane} for a detailed discussion and visualization of the results of HexPlane.}
We address these challenges by adopting a robust 4D grid feature encoding model {that is} capable of representing spatio-temporal information. This model predicts color and density using features derived from the 4D grid. 

To animate 3D videos from a single image, we propose a static-to-dynamic and coarse-to-fine strategy {that structures} the optimization of the 4D representation into three distinct stages. Initially, we develop a robust static 3D model from the reference image 
{with} 2D image diffusion prior~\cite{LDM} and 3D diffusion prior~\cite{Zero-1-to-3}. This static model serves as the initialization of the dynamic model, from which the 3D video animation emerges. In the second stage, we employ a video diffusion prior~\cite{wang2023modelscope} to generate the motion across various timesteps and camera perspectives. During this stage, the reference image is instrumental in aligning the first frame with the source image.
However, a challenge arises as the object in the 3D video tends to drift away from the reference image over time. This drift is largely attributed to the reference image influencing only the initial frame, while subsequent frames primarily rely on the knowledge gained from the video diffusion model.

Personalized modeling techniques~\cite{dreambooth,TextualInversion} are effective for aligning reference images with diffusion priors. However, these methods cannot be applied to video diffusion model when only a single image instead of a video is provided.
The semantic drift of video diffusion {therefore} seems to be 
{inevitable}. Nevertheless, personalized modeling based on a single image is feasible for image diffusion priors, and a video can be conceptualized as a sequence of consecutive images. Consequently, in the third stage of our approach, we employ frame-level processing with personalized modeling to counteract semantic drift.
Specifically, this stage focuses on refining the details and appearance of the 3D video, while preserving its structure and motion. 
We utilize ControlNet-Tile~\cite{controlnet} diffusion prior 
with the second stage 3D video as a condition. Textual Inversion~\cite{TextualInversion} is employed for personalized modeling. Additionally, the ControlNet-Tile diffusion prior not only compensates for reference information but also enhances the video's resolution, as it can be applied effectively to resized low-resolution images.
Through the three-stage optimization process {that is} powered by robust 2D and 3D diffusion priors, our \textbf{\ours} is capable of generating realistic and diverse 3D videos. Our contributions are summarized as follows:
\begin{itemize}
    \item We introduce \ours, a novel framework for animating a single image into a 3D video, utilizing a 4D grid dynamic NeRF representation.
    \item We propose a static-to-dynamic and coarse-to-fine strategy to optimize the 4D representation, integrating 2D, 3D, and personalized modeling diffusion priors.
    \item We conduct extensive qualitative and quantitative experiments to compare \ours with baselines and state-of-the-art text-to-4D method (MAV3D~\cite{mav3d}), demonstrating the superiority of our method.
\end{itemize}

\section{Related Work}

\noindent\textbf{Dynamic Neural Rendering.}
In our framework, we employ Neural Radiance Fields (NeRFs)~\cite{NeRF} to represent the 4D spatio-temporal scenes. 
{NeRFs~\cite{NeRF} enable the rendering of images from diverse target viewpoints by modeling a 3D scene through a neural network that interprets spatial coordinates.}
The architecture of neural network varies, extending from basic MLPs (Multilayer Perceptrons)~\cite{NeRF,MipNeRF} to more complex voxel grid features~\cite{InstantNGP,sun2022direct}. Regarding dynamic NeRFs for 4D spatio-temporal scenes, one popular approach involves separately learning a canonical field and a deformation field across different network layers. However, this technique faces challenges when dealing with changes in scene topology. Another prevailing method decomposes the spatio-temporal dimension into multiple planes~\cite{kplanes,hexplane,shao2023tensor4d}, but this plane-based approach tends to result in the Janus problem when generating 3D video from a single reference image. To overcome these limitations, inspired by Park~\etal~\cite{4dgrid}, 
{we} leverage a 4D grid model to effectively represent dynamic 3D scenes, thereby facilitating the animation of a single image into a 3D video.

\noindent\textbf{Text-to-3D Generation.}
The evolution of multimodal foundation models~\cite{CLIP,LDM,Imagen} has led to significant advancements in in-the-wild text-to-3D generation. Initial efforts~\cite{DreamFields,PureCLIPNeRF,AvatarCLIP} focused on aligning text prompts with rendered images using CLIP~\cite{CLIP}. Recognizing the detailed semantic capabilities of diffusion models, DreamFusion~\cite{DreamFusion} and SJC~\cite{SJC} introduced techniques to distill knowledge from text-based diffusion models into 3D representations, yielding promising outcomes. Subsequently, more advanced methods~\cite{Magic3D,wang2023prolificdreamer,Fantasia3D} have emerged to further refine diffusion-based 3D generation, incorporating mesh representation and distribution optimization. In the realm of dynamic scenes, MAV3D~\cite{mav3d} has been developed %
{to optimize} a dynamic NeRF representation through a static-to-dynamic strategy.

\noindent\textbf{Image-to-3D Generation.}
Lifting a single image to 3D assets plays a crucial role in 3D content creation. Initial research in this area focused on domain-specific 3D lifting, employing strong domain priors for specific subjects like human bodies~\cite{SMPL,3D-pose-baseline} and faces~\cite{3DMM,3DMM-learnt}. Recently, the emergence of potent diffusion priors has enabled the realization of in-the-wild image-to-3D generation. One prominent approach~\cite{Zero-1-to-3,liu2023syncdreamer,one-2-3-45} involves fine-tuning 2D image generation models to produce multi-view images directly. Alternatively, some methods~\cite{RealFusion,magic123,tang2023dreamgaussian} combine 2D diffusion with the above fine-tuned multi-view 3D diffusion to optimize 3D representation by distilling diffusion knowledge. However, these techniques primarily focus on static scenes and do not incorporate aspects of animation.

\noindent\textbf{Animating Single Image to Video.}
Image-to-video generation has gained considerable popularity in both academic circles~\cite{luo2023videofusion,chen2023videocrafter1,wang2023videocomposer} and commercial applications\footnote{https://www.pika.art/}\footnote{https://research.runwayml.com/gen2}. These methods typically utilize the reference image either as the initial frame or to extract semantic information as a condition. 
{We recognize the potential of animating the well-established static 3D scene with text, especially for applications in the metaverse and video game industries. 
This motivates us to pioneer the exploration of in-the-wild image-to-3D-video generation.}

\begin{figure*}[t]
    \centering
    \includegraphics[width=.99\textwidth]{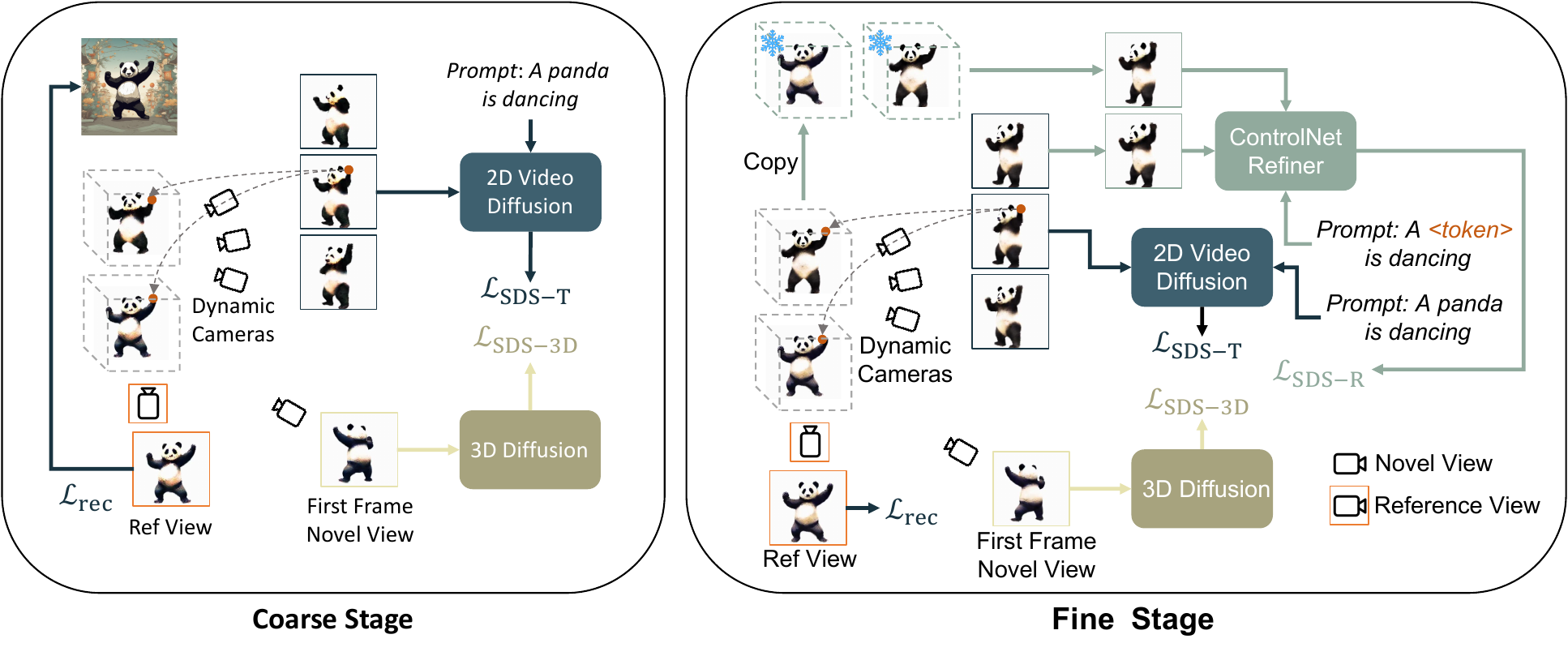}
    \vspace{-.15in}
    \caption{The overall framework of {our} \ours. After learning the static scene (the first stage, not shown in the figure), the dynamic scene is optimized with a coarse-to-fine strategy in two stages. In the coarse stage, we optimize the dynamic NeRF with the combination of video diffusion and 3D diffusion priors. Subsequently, in the fine stage, additional ControlNet prior is introduced to refine the details and correct semantic drift. The condition of ControlNet derives from the frozen coarse stage model to reduce error accumulation.}
    \label{fig:framework}
\end{figure*}

\section{Our Method}

\noindent\textbf{Overview.}
{Our framework leverages} static-to-dynamic and coarse-to-fine strategy to animate a single image into a 3D video, structured in three distinct stages. 
{Specifically, we develop a static NeRF using the reference image in the first stage (static-stage) under the guidance of both 2D and 3D diffusion priors. }
{Subsequently in} the second stage (coarse-stage), the 4D grid dynamic NeRF is initialized from the static NeRF and further refined with the assistance of video diffusion prior and 3D diffusion prior. 
{In the final stage (fine-stage), we employ a personalized image diffusion prior to mitigate the semantic drift introduced by the video diffusion model. The image diffusion prior is specifically fine-tuned with the reference image to provide additional supervision.}
The comprehensive framework encompassing both the coarse and fine stages is illustrated in Fig.~\ref{fig:framework} 
with the static optimization stage excluded due to space constraints.

\subsection{4D Grid Encoding}
Given a camera position and a timestep $t$, a ray is cast from the camera 
center through each pixel on the image 
penetrating into the scene. 
{We sample 3D points along this ray and determine the color and density to render the image through volume rendering.}
Multi-scale grid encoding~\cite{InstantNGP} is an efficient and effective method for storing and representing 3D scenes. The features extracted from this grid are instrumental in calculating both density and color. 
In the temporal domain, Fridovich-Keil~\etal~\cite{kplanes} have demonstrated that multi-scale grid encoding is not essential. 
Consequently, we construct our 4D grid in the following manner: we divide the time dimension evenly into $T$ grids, and for each time grid, we establish a 3D multi-scale grid $V$ (excluding hash encoding). Spatio-temporal features $F_{x,y,z,t}$ are then linearly interpolated from the two nearest time grids:
\begin{equation}
    F_{x,y,z,t} = \frac{t_{+} -t}{\Delta_t} V_{x,y,z, t_{-}} + \frac{t - t_{-}}{\Delta_t} V_{x,y,z, t_{+}},
\end{equation}
where $x$,$y$ and $z$ represent the spatial coordinates, 
{and} $t$ denotes the normalized time within the range of $[0,1]$. $t_{+}$, $t_{-}$ and $\Delta_t$ refer to the nearest upper time grid, lower time grid and the time interval between two consecutive grids. Utilizing these spatio-temporal features, we are able to generate color $c$ and density $\tau$ via projection MLPs. In line with DreamFusion~\cite{DreamFusion}, we generate albedo and %
{simulate} random light sources to accurately represent color.

\noindent\textbf{Temporal Total Variation Loss.}
To effectively transmit the information from the first frame to subsequent time grids while promoting temporal smoothness, we apply a total variation (TV) loss~\cite{tvloss} to the 3D grid $V$ across the adjacent time dimensions:
\begin{equation}
\label{eq:tv}
    \mathcal{L}_{TV} = \sum_{t=0}^{T-1} \sum_{x,y,z} (V_{x,y,z,t} - V_{x,y,z,t+1})^2.
\end{equation}

\subsection{Static Scene Optimization}
Following Magic123~\cite{magic123}, we employ Score Distillation Sampling (SDS) losses from both 2D and 3D diffusion priors to guide the optimization of the static scene. This strategy effectively enhances both texture quality and 3D geometry.
Specifically, stable diffusion~\cite{LDM} conditioned on the text prompt $e$ is adopted as 2D diffusion prior and Zero-1-to-3-XL~\cite{Zero-1-to-3} conditioned on the reference image $\tilde{\mathbf{I}}^r$ and relative pose $\Delta p$ is adopted as 3D diffusion prior. The SDS loss is formulated as:
\begin{equation}
\small
\begin{aligned}
    \mathcal{L}_{\mathrm{SDS}} &= \mathbb{E}_{\sigma, p, \epsilon}\left[ \omega(\sigma) \left(\epsilon_\phi^{2D}\left(\mathbf{I}^p ; \sigma, e\right)-\epsilon\right) \frac{\partial \mathbf{I}^p}{\partial \theta_s}\right] \\
    &+ \lambda_{3D} \mathbb{E}_{\sigma, p, \epsilon}\left[ \omega(\sigma) \left(\epsilon_\phi^{3D}\left(\mathbf{I}^p ; \sigma, \tilde{\mathbf{I}}^r, \Delta p\right)-\epsilon\right) \frac{\partial \mathbf{I}^p}{\partial \theta_s}\right],
\end{aligned}
\end{equation}
where $\theta_s$ represents the parameters of the static NeRF model and $\mathbf{I}^p$ is the rendered RGB image from the camera position $p$. $\epsilon_\phi^{2D}(\cdot)$ and $\epsilon_\phi^{3D}(\cdot)$ denote the predicted noise by 2D and 3D diffusion priors, respectively. $\omega(\sigma)$ refers to a weighting function corresponding to the noise timestep $\sigma$.

Additionally, we leverage the RGB $\tilde{\mathbf{I}}^r$, foreground mask $\tilde{M}^r$ and depth $\tilde{d}^r$ from the reference image to further refine the model from the reference view:
\begin{equation}
\label{eq:rec}
\begin{aligned}    
    \mathcal{L}_{\mathrm{rec}} &= \lambda_{rgb} ||\tilde{M}^r \odot (\tilde{\mathbf{I}}^r - {\mathbf{I}}^r)|| + \lambda_{mask} ||\tilde{M}^r - {M}^r|| \\
    &+ \lambda_{d} \left[1-\frac{\operatorname{cov}\left(\tilde{M}^r \odot \tilde{d}^r, \tilde{M}^r \odot d^r \right)}{\operatorname{std}\left(\tilde{M}^r \odot \tilde{d}^r\right) \operatorname{std}(\tilde{M}^r \odot d^r)}\right],
\end{aligned}
\end{equation}
where $\lambda_{rgb}$, $\lambda_{mask}$ and $\lambda_{d}$ denote the weights of RGB, mask and depth loss. $\odot$ denotes Hadamard product. $\operatorname{cov}$ and $\operatorname{std}$ denote covariance and standard deviation, respectively. The static scene is optimized by the combination of the above two losses:
\begin{equation}
    \mathcal{L}_{\mathrm{static}} = \mathcal{L}_{\mathrm{SDS}} + \mathcal{L}_{\mathrm{rec}}.
\end{equation}

\subsection{Coarse Dynamic Scene Optimization}
The dynamic NeRF is initialized from the static NeRF. Specifically, each 3D multi-scale grid $V$ within the $T$ time grids is initialized with the parameters of the static model. All the time grids share the same projection layers that were pre-trained in the static stage. 
The dynamic model is {thus} initialized as a model that can generate 3D video where each frame is the same static scene.  
We {then} optimize it to align with the motion described by the text prompt $e$. We distill latent video diffusion model~\cite{wang2023modelscope} with SDS loss~\cite{DreamFusion,mav3d} to achieve this goal. Specifically, given the camera trajectory $r(t)$ of a video with a fixed number of frames $N_f$, we can cast rays and sample timesteps based on the frame rate. 
{Subsequently,} a video $\mathbf{V}^{r(t)}$ is rendered from the dynamic NeRF and fed into the video diffusion prior to for SDS loss:
\begin{equation}
\small
\label{eq:v-sds}
    \mathcal{L}_{\mathrm{SDS-T}} = \mathbb{E}_{\sigma, r(t), \epsilon}\left[ \omega(\sigma) \left(\epsilon_\phi^{Vid}\left(\mathbf{V}^{r(t)} ; \sigma, e\right)-\epsilon\right) \frac{\mathbf{V}^{r(t)}}{\partial \theta_d}\right],
\end{equation}
where $\theta_d$ is the model parameters and $\epsilon_\phi^{Vid}(\cdot)$ denotes the predicted noise by the video diffusion prior.

\begin{figure}[t]
    \centering
    \begin{subfigure}[c]{0.49\linewidth} 
        \centering
        \includegraphics[width=\textwidth]{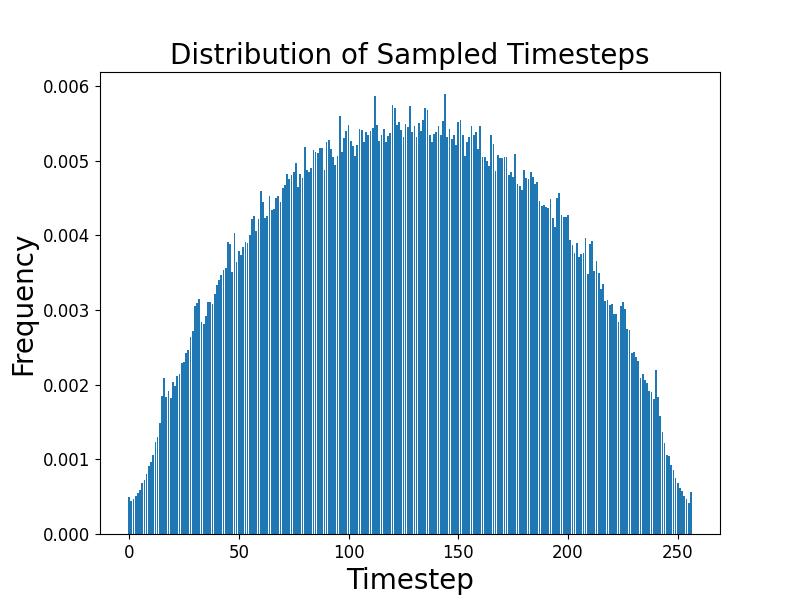}
        \caption{Random Sampling.}
        \label{fig:random-sampling}
    \end{subfigure}
    \hfill 
    \begin{subfigure}[c]{0.49\linewidth} 
        \centering
        \includegraphics[width=\textwidth]{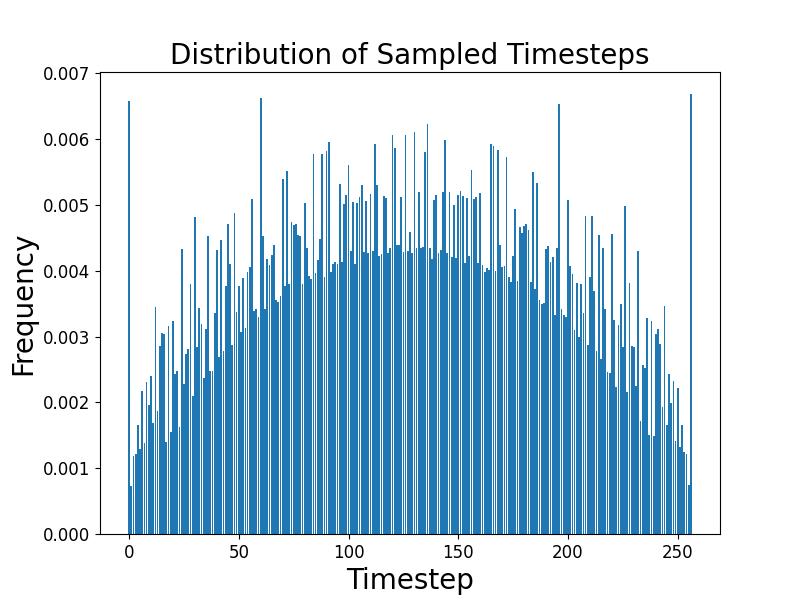}
        \caption{Temporal Balanced Sampling.}
        \label{fig:balanced-sampling}
    \end{subfigure}
    \vspace{-.1in}
    \caption{Distribution of the sampled timestep.}
    \vspace{-.1in}
\end{figure}

\noindent\textbf{Temporal Balanced Sampling.}
Unlike static 3D scenes, dynamic scenes necessitate sampling frames over a temporal range of $t\in [0,1]$. MAV3D utilizes Make-A-Video~\cite{mav} as its diffusion prior, which can condition on the video frame rate. However, this model is not publicly available. Consequently, we adopt ModelScope~\cite{wang2023modelscope} as our video diffusion prior. ModelScope is not trained for extremely high frame rates and also lacks the capability to condition on the frame rate. To mitigate issues related to extreme frame rates, we limit the FPS (frames per second) to a range of $[16, 256]$. 
{At each iteration, we sample $N_f$ timesteps from a randomly chosen starting timestep and FPS.
}
The distribution of sampled timesteps is depicted in Fig.~\ref{fig:random-sampling}.
However, random sampling tends to result in the beginning and ending timesteps being less frequently sampled compared to the middle timesteps, leading to suboptimal optimization of the first and last time grids (examples are provided in Appendix.~\ref{sec:sample}). Moreover, as the reference image constitutes the first frame of the generated video, it is beneficial to sample more from the first frame to retain the reference information.  
{As a result,} we allocate a higher probability of $\alpha$ specifically for sampling timesteps that begin at time 0 (the first frame), and similarly, a probability of $\alpha$ for timesteps concluding at time 1.
The distribution of this temporal balanced sampling method is illustrated in Fig.~\ref{fig:balanced-sampling}.

\noindent\textbf{First Frame Supervision.}
Since we animate the reference image to a 3D video, the reference image, {which serves as the first frame, offers additional supervision for the video generation process.}
{Specifically, we enhance the supervision of the dynamic NeRF by incorporating both the reconstruction loss $\mathcal{L}_{\mathrm{rec}}$ (as detailed in Eq.~\ref{eq:rec}) and the 3D diffusion prior SDS loss $\mathcal{L}_{\mathrm{SDS-3D}}$:
}
\begin{equation}
\small
\label{eq:3d-sds}
    \mathcal{L}_{\mathrm{SDS-3D}} = \mathbb{E}_{\sigma, p, \epsilon}\left[ \omega(\sigma) \left(\epsilon_\phi^{3D}\left(\mathbf{V}^{r(t)}_0 ; \sigma, \tilde{\mathbf{I}}^r, \Delta p\right)-\epsilon\right) \frac{\partial \mathbf{V}^{r(t)}_0}{\partial \theta_d}\right],
\end{equation}
{only on the first frames of the sampled videos $\mathbf{V}^{r(t)}_0$, which occurs with an approximate probability of $\alpha$.}
The final loss is {thus} formulated as:
\begin{equation}
\begin{aligned}
\mathcal{L}_{\mathrm{dynamic}} &= \mathcal{L}_{\mathrm{SDS-T}} + \lambda_{TV} \mathcal{L}_{\mathrm{TV}} \\
&+ \mathbbm{1}_{t_0 = 0} (\mathcal{L}_{\mathrm{rec}} + \lambda_{3D} \mathcal{L}_{\mathrm{SDS-3D}} ),
\end{aligned}
\end{equation}
where $\mathcal{L}_{\mathrm{TV}}$ denotes the total variation loss (Eq.~\ref{eq:tv}) and $\lambda_{TV}$ is the weight for this loss.

\subsection{Semantic Refinement}
During the coarse dynamic scene optimization stage, 
information from the reference image is exclusively applied to the first frame. 
Details of the objects in subsequent frames are derived solely from the video diffusion prior 
{and} guided by the text prompt $e$. For {example}, in a scenario where a ``panda'' is animated to dance, frames beyond the first are likely to distill a ``panda'' representation from the video diffusion prior that is potentially different from the reference image. Consequently, semantic drift becomes inevitable in the coarse stage.

In image-to-3D generation, personalized modeling~\cite{TextualInversion,dreambooth} is commonly used to represent the reference image in the text-to-image model.
RealFusion~\cite{RealFusion} utilizes a unique token learned through textual inversion~\cite{TextualInversion} to represent the reference image in the text-to-image diffusion prior. DreamCraft3D~\cite{sun2023dreamcraft3d} optimizes the text-to-image diffusion prior using augmented renderings of the reference image, as facilitated by DreamBooth~\cite{dreambooth}. However, a single image is insufficient to learn a personalized model or token for a text-to-video diffusion prior.
As a solution, we approach each frame independently and optimize it using the personalized text-to-image diffusion prior. Utilizing a naive text-to-image (T2I) personalized model 
{can} introduce unexpected motion changes 
{since} the T2I model lacks awareness of the other frames. Our objective is to have the T2I model concentrated solely on refining textures and details. To this end, ControlNet-Tile~\cite{controlnet} is an ideal diffusion prior 
as it conditions on a low-resolution image while refining details and enhancing resolution.
Accordingly, we learn a token to represent the reference image for the base model of ControlNet (Stable Diffusion v1.5) and optimize individual frames ${\mathbf{I}}_t^{r(t)}$ with this token. To prevent error accumulation, the conditioning image $\hat{\mathbf{I}}_t^{r(t)}$ is generated from the fixed dynamic NeRF model established in the coarse stage. This diffusion prior also guides the dynamic NeRF through SDS loss, which is formulated as follows:
\begin{equation}
\small
    \mathcal{L}_{\mathrm{SDS-R}} = \mathbb{E}_{\sigma, r(t), \epsilon}\left[ \omega(\sigma) \left(\epsilon_\phi^{CN}\left({\mathbf{I}}_t^{r(t)} ; \sigma, \hat{\mathbf{I}}_t^{r(t)}, e\right)-\epsilon\right) \frac{\mathbf{I}_t^{r(t)}}{\partial \theta_d}\right],
\end{equation}
where $\theta_d$ is the parameters of the dynamic NeRF model and $\epsilon_\phi^{CN}(\cdot)$ denotes the predicted noise by the personalized ControlNet diffusion prior. The video and 3D diffusion priors are also leveraged in this stage, so the final loss is:
\begin{equation}
\begin{aligned}
\mathcal{L}_{\mathrm{refine}} &= \mathcal{L}_{\mathrm{SDS-T}} + \lambda_R \mathcal{L}_{\mathrm{SDS-R}} + \lambda_{TV} \mathcal{L}_{\mathrm{TV}} \\
&+ \mathbbm{1}_{t_0 = 0} (\mathcal{L}_{\mathrm{rec}} + \lambda_{3D} \mathcal{L}_{\mathrm{SDS-3D}} ),
\end{aligned}
\end{equation}
where $\mathcal{L}_{\mathrm{SDS-T}}$, $\mathcal{L}_{\mathrm{TV}}$, $\mathcal{L}_{\mathrm{rec}}$ and $\mathcal{L}_{\mathrm{SDS-3D}}$ denote the video diffusion SDS loss (Eq.~\ref{eq:v-sds}), total variation loss (Eq.~\ref{eq:tv}), first frame reconstruction loss (Eq.~\ref{eq:rec}) and first frame 3D diffusion SDS loss (Eq.~\ref{eq:3d-sds}), respectively. $\lambda_{R}$, $\lambda_{TV}$ and $\lambda_{3D}$ are the weights for ControlNet SDS loss, TV loss and 3D prior SDS loss, respectively.

\begin{figure*}[t]
    \centering
    \includegraphics[width=.95\textwidth]{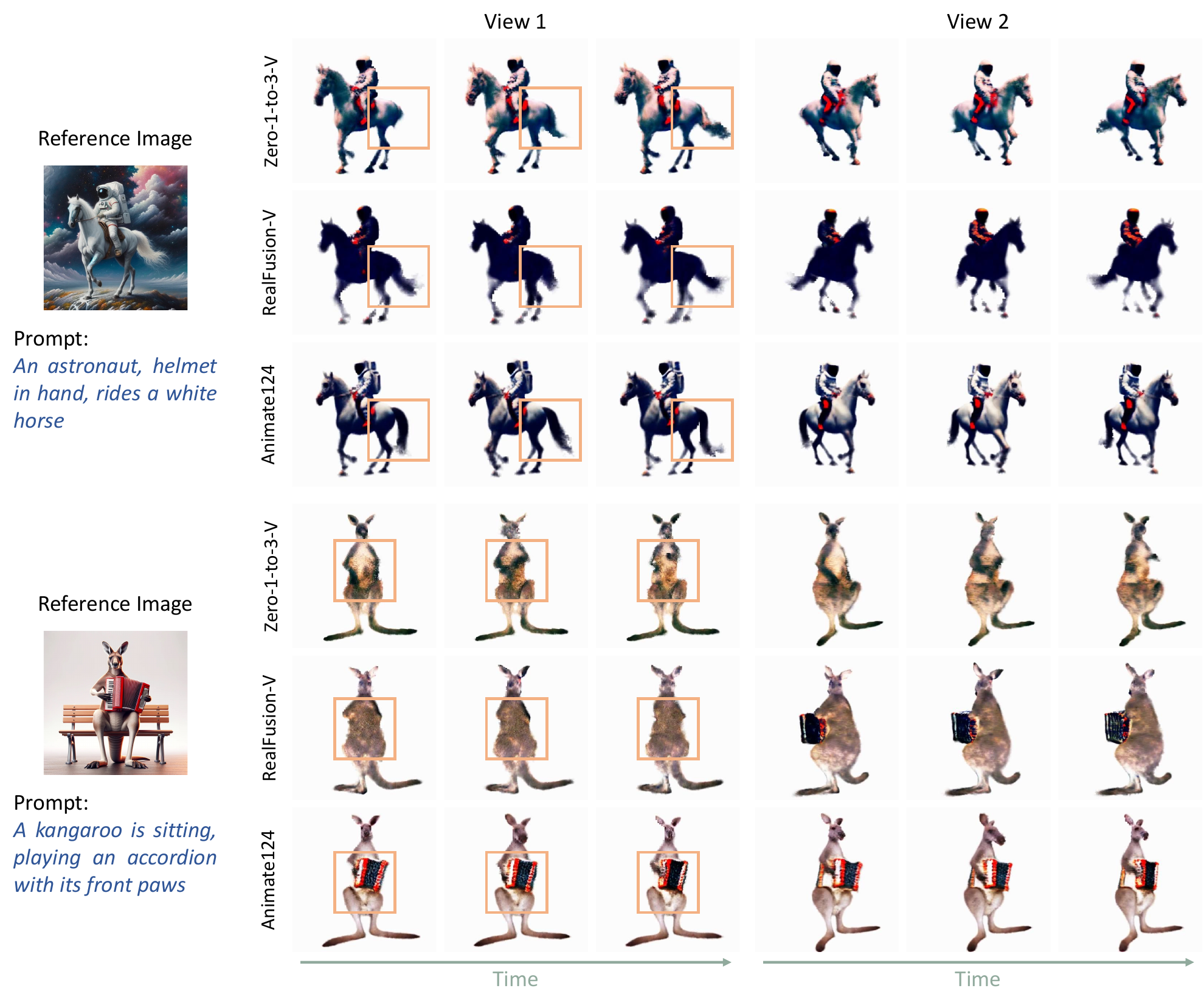}
    \vspace{-.2in}
    \caption{Qualitative comparison with the baseline methods on image-text-to-4D generation. Results in two views are shown. View 1 is the reference view and view 2 is another view. We use \textcolor{mycolor}{square} in view 1 to better illustrate the motion and difference among methods.}
    \label{fig:sota}
\end{figure*}

\noindent\textbf{Over-Saturation and Over-Smoothing.}
Score Distillation Sampling (SDS)~\cite{DreamFusion} requires a large classifier-free guidance (CFG) scale to effectively distill knowledge from text-to-image diffusion models. This requirement arises because a large CFG scale can diminish the diversity of the T2I model, focusing more on fidelity to the given text. However, as indicated by qualitative results from DreamFusion~\cite{DreamFusion} and ProlificDreamer~\cite{wang2023prolificdreamer}, SDS often encounters issues of over-saturation and over-smoothing due to the heightened CFG scale. Our coarse stage model faces similar challenges.
ControlNet-Tile~\cite{controlnet} conditions on a low-resolution coarse image, mirroring the effect of ancestral sampling~\cite{DDPM,dpm-solver}. This effect allows us to use a CFG scale comparable to that in standard image generation tasks (\eg, a scale of 7.5) in this diffusion prior, which can mitigate the over-saturation and over-smoothing problems observed in our coarse model.

\begin{figure*}
    \begin{subfigure}[c]{\textwidth}
    \centering
    \includegraphics[trim=0cm 10cm 0cm 0cm, clip, width=.95\textwidth]{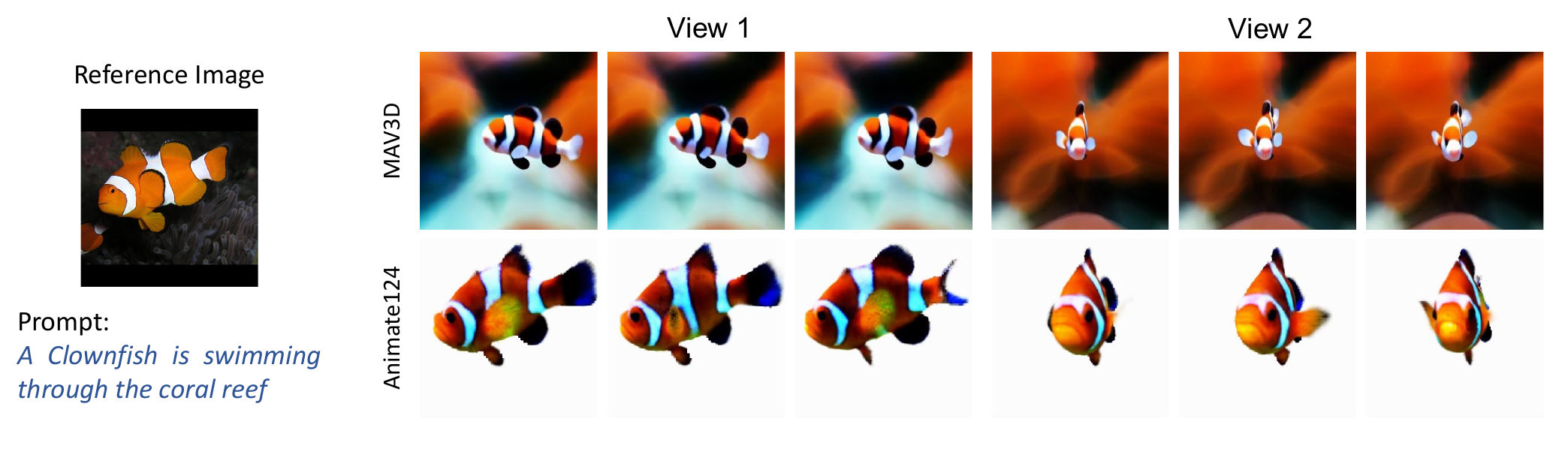}
    \caption{Comparison with MAV3D text-to-4D generation. Reference image is not used in MAV3D. \ours can generate dynamic motion with the reference image as the protagonist while MAV3D cannot precisely control the subject.}
    \label{fig:mav3d-text}
    \end{subfigure} \\
    \begin{subfigure}[c]{\textwidth}
    \centering
    \includegraphics[trim=0cm 0cm 0cm 10cm, clip, width=.95\textwidth]{figures/mav3d.pdf}
    \caption{Comparison with MAV3D image-to-4D generation. Text prompt is not used in MAV3D. \ours can better preserve the appearance and pose of the reference image.}
    \label{fig:mav3d-image}
    \end{subfigure}
    \vspace{-.1in}
\caption{Qualitative comparison with MAV3D on 4D generation with the samples on its website. \textbf{Note} that we extract frames from the videos on MAV3D website, 
{thus} the images in two views may not be perfectly matched.}
\label{fig:mav3d}
\end{figure*}

\section{Experiment}

\subsection{Experimental Setup}

\noindent\textbf{Implementation Details.}
In the static stage, static NeRF model is optimized for 5,000 iterations. 
The model contains 16 levels of grid encoding, with resolutions ranging from 16 to 2,048, and a dimension of 2 for each level. Two layers of MLPs with 64 hidden dimensions are used to calculate density and albedo.
For the reference view reconstruction loss, $\lambda_{rgb}$, $\lambda_{mask}$ and $\lambda_{d}$ are set to 5, 0.5 and 0.001, respectively. Stable Diffusion v1.5~\cite{LDM} and Zero-1-to-3-XL~\cite{Zero-1-to-3} serve as our 2D and 3D diffusion priors for this stage 
with $\lambda_{3D}$ set to 40.
In the coarse dynamic scene optimization stage, dynamic NeRF is optimized for 10,000 iterations with the same loss weight as in the static stage. The time grid size of dynamic NeRF is set to 64, and $\lambda_{TV}$ is set to 0.1 for regularization purposes.
In the semantic refinement stage, the model is further trained for 5,000 iterations and the loss weight for ControlNet SDS loss, $\lambda_{R}$ is adjusted to 1.
Adam~\cite{kingma2014adam} with a learning rate 0.001 is adopted across all stages, and the rendering resolution is 128$\times$128.

\noindent\textbf{Camera Setting.} 
As for the reference view, following Magic123~\cite{magic123}, we assume the reference image is shot from the front view (polar angle $90^\circ$ and azimuth angle $0^\circ$) with the radius 1.8 meters, and the field of view (FOV) of the camera is set to $40^\circ$.
During dynamic training stages, we adopt dynamic camera~\cite{mav3d} to simulate the camera motion in the text-to-video diffusion model.

\begin{table*}[ht]
    \caption{Comparison with other methods. Across all metrics, higher scores indicate better results. \textbf{Human evaluation is shown as a percentage of majority votes favoring the baseline compared to our model in the specific setting}.}
    \vspace{-.1in}
    \label{tab:comparison}
    \centering
    \small
    \resizebox{\linewidth}{!}{%
    \begin{tabular}{l|cccc|ccccc}
    \toprule
    \multirow{2}{*}{Model} & \multicolumn{4}{c|}{CLIP Evaluation} & \multicolumn{5}{c}{Human Evaluation} \\
    & \multirow{1}{*}{CLIP-R} & \multirow{1}{*}{CLIP-T} &  \multirow{1}{*}{CLIP-I} & \multirow{1}{*}{CLIP-F} & Text Align. & Image Align. & Video Quality & Realistic Motion & More Motion \\
    \midrule
    Zero-1-to-3-V~\cite{Zero-1-to-3}  & 0.8746 & 0.3004 & 0.7925 & 0.9663 & 0.1667 & 0.1833 & 0.1833 & 0.1750 & 0.2667 \\
    RealFusion-V~\cite{RealFusion}  & 0.8939 & 0.3075 & 0.8026  & 0.9691 & 0.2583 & 0.3042 & 0.2750 & 0.2625 & 0.4083 \\
    \textbf{\ours}  & \textbf{0.9311} & \textbf{0.3170} & \textbf{0.8544} & \textbf{0.9781} & --- & --- & --- & --- & --- \\
    \bottomrule
    \end{tabular}
    }
\end{table*}

\begin{table}[ht]
    \caption{Ablation studies on the proposed components.}
    \vspace{-.1in}
    \label{tab:ablation}
    \centering
    \small
    \begin{tabular}{l|cccc}
    \toprule
    \multirow{1}{*}{Model} & \multirow{1}{*}{CLIP-R}  & \multirow{1}{*}{CLIP-T} &  \multirow{1}{*}{CLIP-I} & \multirow{1}{*}{CLIP-F} \\
    \midrule
    w.o. 3D Prior  & 0.9042 & 0.3020 & 0.8141 & 0.9704 \\
    w.o. BalancedSampl. & \textbf{0.9377} & \textbf{0.3205} & 0.8083 & 0.9728 \\
    w.o. SemRefine.  & 0.9221 & 0.3129 & 0.8331 & 0.9715 \\
    \midrule
    \textbf{\ours} & {0.9311} & {0.3170} & \textbf{0.8544} & \textbf{0.9781} \\
    \bottomrule
    \end{tabular}
\end{table}

\noindent\textbf{Benchmark and Evaluation Metrics.}
As the first endeavor to animate a single image into a 3D video, we build a benchmark comprising 24 image-text pairs for evaluation. 
Our methodology is assessed across three dimensions: text-video alignment, image-video alignment, and overall video quality. For each 3D video, we render views from 10 different views around the scene.
In terms of text-video alignment, we measure the retrieval accuracy of text prompts (CLIP-R~\cite{DreamFields}) and compute the image-text cosine similarity (CLIP-T) for every frame of the rendered videos. For image-video alignment, we render a video from the reference camera pose and then calculate the cosine similarity between the CLIP visual features of each frame and the reference image.
Regarding video quality, we evaluate the consistency between frames by calculating the cosine similarity between CLIP visual features of every two consecutive frames in each rendered video. For these assessments, we utilize the CLIP~\cite{CLIP} ViT-B/32 variant.
In addition, following MAV3D~\cite{mav3d}, we conduct user studies on five qualitative metrics: (1) similarity to the reference image; (2) faithfulness to the textual prompt; (3) video quality; (4) realism of motion; and (5) amount of motion.

\subsection{Comparison with Other Methods}
{As the pioneering work on} image-text-to-4D generation, we have developed two baselines for comparative analysis. These baselines employ distinct image-to-3D static generation methods to establish a static NeRF, which is subsequently optimized using SDS loss derived from a video diffusion prior. Specifically, our baselines: Zero-1-to-3-V and RealFusion-V utilize Zero-1-to-3~\cite{Zero-1-to-3} and RealFusion~\cite{RealFusion} in their static stages, respectively.
{Furthermore, we conduct a qualitative comparison of \ours with MAV3D to provide a comprehensive evaluation.}
This comparison focuses on the prompts and examples featured on the MAV3D website\footnote{https://make-a-video3d.github.io/}, thus allowing for a detailed analysis of our approach in relation to existing methodologies.

\noindent\textbf{Comparison with Baselines.}
In our comparative analysis, \ours is evaluated against two baselines, Zero-1-to-3-V~\cite{Zero-1-to-3} and RealFusion-V~\cite{RealFusion}, as presented in Tab.~\ref{tab:comparison} and Fig.~\ref{fig:sota}. As reported in Tab.~\ref{tab:comparison}, \ours outperforms both baselines 
{quantitatively} in terms of CLIP and human evaluations. Fig.~\ref{fig:sota} illustrates that Zero-1-to-3-V~\cite{Zero-1-to-3} fails to preserve the original appearance of the reference image. For 
example, the first example exhibits a color shift to red, and the ``accordion'' in the second example is absent. RealFusion-V~\cite{RealFusion}, on the other hand, exhibits inconsistencies in 3D geometry. As shown in the second example, ``View 1'' should represent the reference view, but the ``kangaroo'' is inaccurately rotated. In contrast, \ours successfully maintains both consistent 3D geometry and the fidelity of the reference image appearance.

\noindent\textbf{Comparison with MAV3D.}
We compare \ours with MAV3D~\cite{mav3d} on two distinct settings, as illustrated in Fig.~\ref{fig:mav3d}.
{The text-to-4D generation of MAV3D} relies solely on text prompts, while its image-to-4D generation exclusively utilizes the reference image as a prompt. In contrast, \ours leverages both the reference image and the text prompt. To facilitate a more direct comparison with {the image-to-4D approach of MAV3D}, we refrain from specifying motion through text in Fig.~\ref{fig:mav3d-image} {and use the textual name of the object as the prompt instead.} %
Fig.~\ref{fig:mav3d-text} demonstrates that \ours is capable of generating dynamic motion that aligns the protagonist with the reference image. In comparison, MAV3D (Fig.~\ref{fig:mav3d-text}) struggles to control the protagonist in the 3D video. Regarding image-to-4D generation, MAV3D fails to preserve the original appearance of the reference image, as evidenced by the flamingo's body turning black. Conversely, \ours successfully produces more realistic videos 
{and} consistently maintaining the appearance of the reference image. These outcomes highlight the efficacy of our method.

\subsection{Ablation Studies and Analysis}

\noindent\textbf{Effectiveness of Semantic Refinement.}
Semantic refinement aims at alleviating semantic drift of the video generation model with the personalized ControlNet~\cite{controlnet}. As shown in {the} 3rd and last row of Tab.~\ref{tab:ablation}, semantic refinement can improve the performance on text alignment, image alignment, and video consistency. The improvement in the image alignment is the most significant, further demonstrating the effectiveness of addressing semantic drift.
Qualitative evaluation and the comparison with super-resolution prior are illustrated in Appendix.~\ref{sec:refine}.

\noindent\textbf{Effectiveness of 3D Diffusion Prior.}
3D diffusion prior serves as a strong supervision for the first frame, which helps to learn both geometry and texture in all three stages. 
In the first and last row of Tab.~\ref{tab:ablation}, removing the 3D prior impairs the overall geometry and thus influences all three aspects of the generated video.

\noindent\textbf{Effectiveness of Temporal Balanced Sampling.}
Removing temporal balanced sampling leads to the first frame supervision being almost ignored in the dynamic stages. The model degenerates to text-to-4D generation with image-to-3D static initialization and semantic refinement. 
Therefore, this model has good text alignment but the similarity to the reference image is quite poor. 
In addition, this model is not consistent in 3D geometry with severe Janus problem due to the lack of guidance from the 3D diffusion prior.

\begin{figure*}
    \centering
    \includegraphics[width=.99\linewidth]{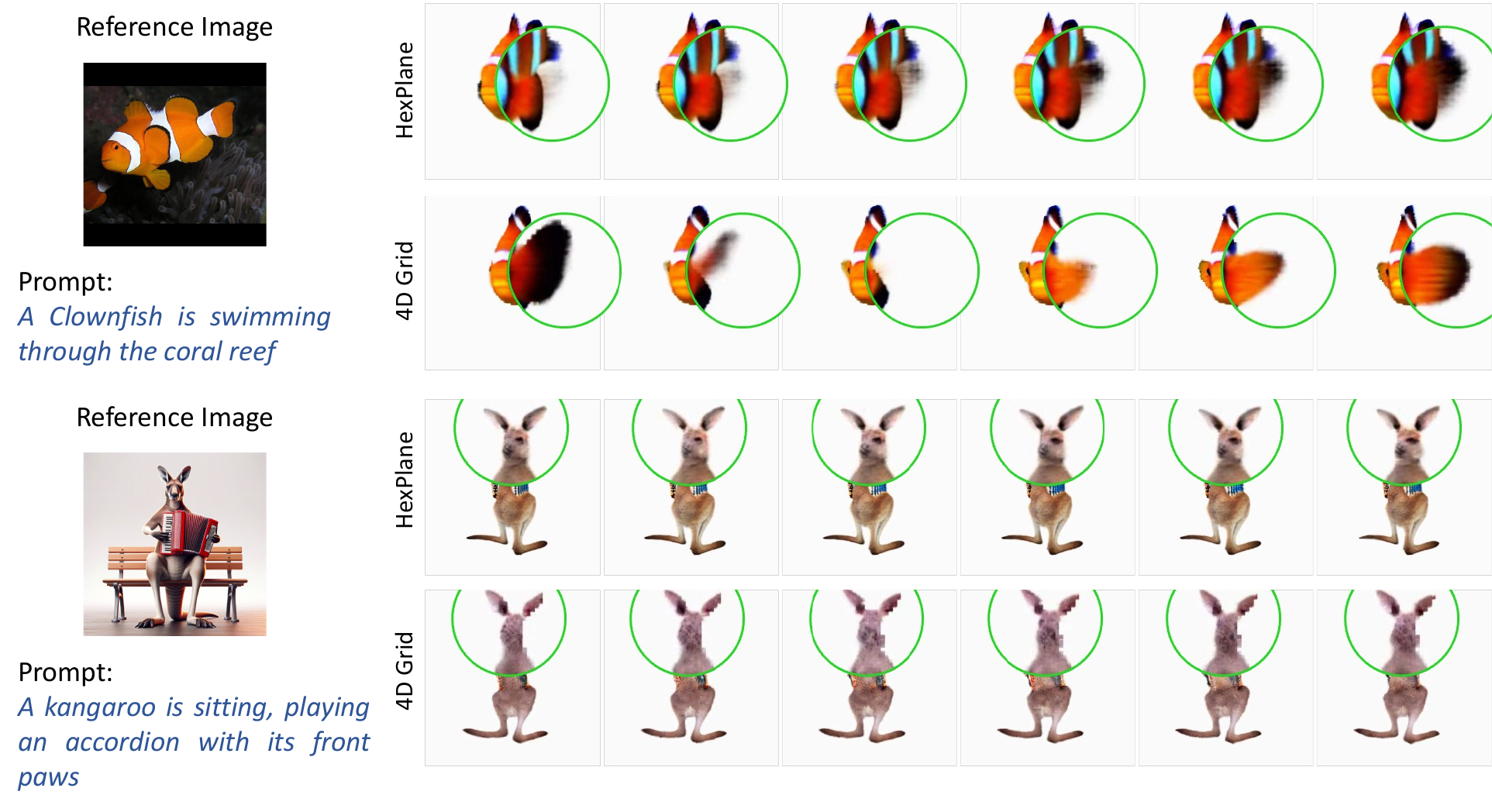}
    \caption{The comparison between 4D grid encoding and HexPlane.  In each example, six consecutive frames are displayed. HexPlane is observed to struggle with limited motion and a conspicuous Janus problem.}
    \label{fig:hexplane}
\end{figure*}

\section{Conclusion}
We introduce \ours, the first work to animate a single in-the-wild image into a 3D video through textual motion descriptions. With the guidance of three diffusion priors, we optimize an advanced 4D grid dynamic NeRF in a static-to-dynamic and coarse-to-fine manner. Specifically, we leverage 2D image and 3D diffusion priors to develop a static 3D scene, which is then animated with video diffusion prior. To address the semantic drift inherent in the video diffusion prior, we further proposed semantic refinement, incorporating a personalized diffusion prior as additional supervision.
With the innovative three-stage framework, \ours is capable of producing high-quality 4D dynamic scenes from both the reference image and textual descriptions.

\appendix

\begin{figure*}
    \centering
    \includegraphics[width=.99\linewidth]{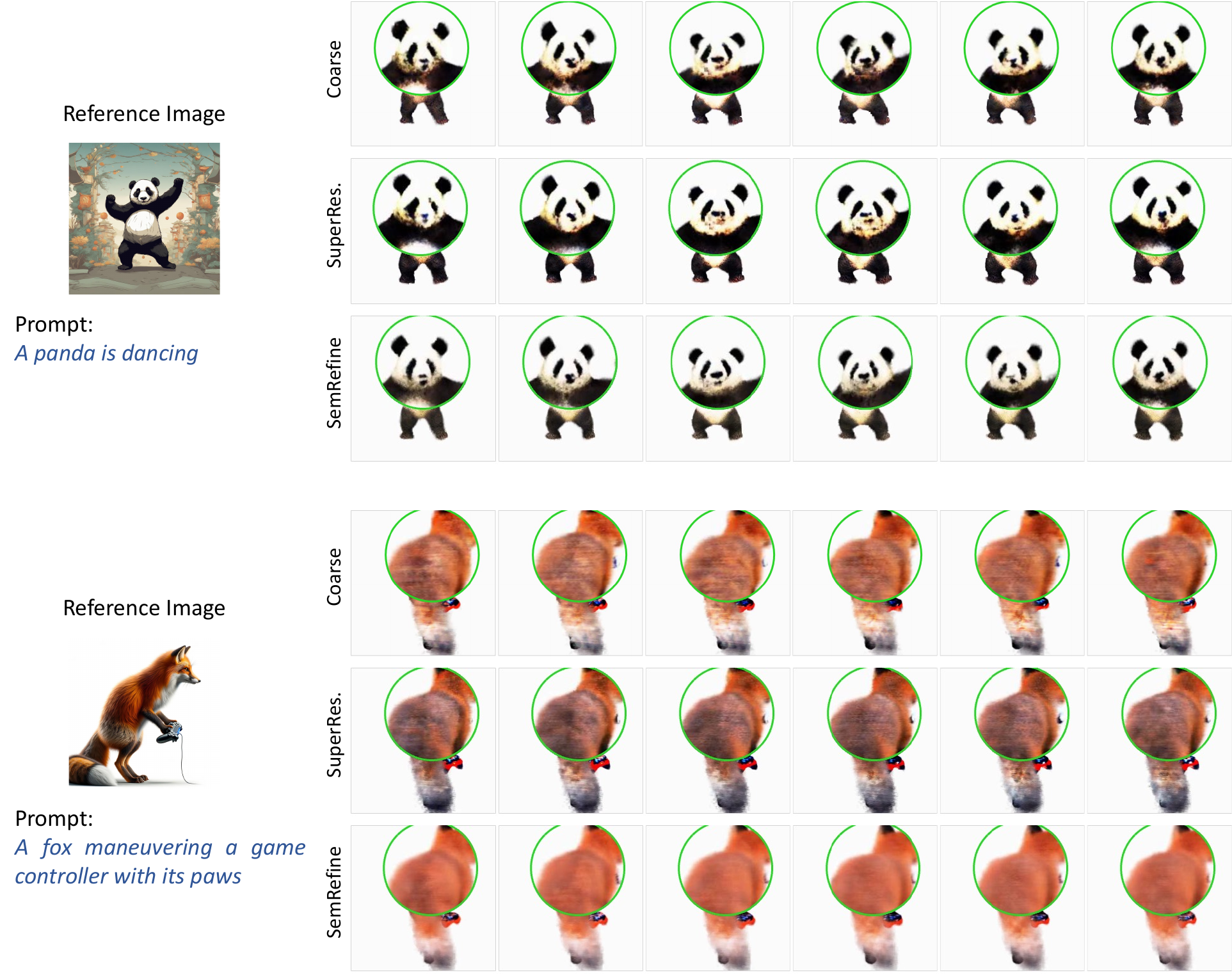}
    \caption{The comparison between semantic refinement and super-resolution refinement in the dynamic fine stage. In each example, the first row depicts the 3D video as generated in the coarse stage, while the second and third rows show the outcomes following semantic and super-resolution refinement, respectively. Semantic refinement yields superior results by incorporating more semantic information of the reference image.}
    \label{fig:sr}
\end{figure*}

\begin{figure*}
    \centering
    \includegraphics[width=.99\linewidth]{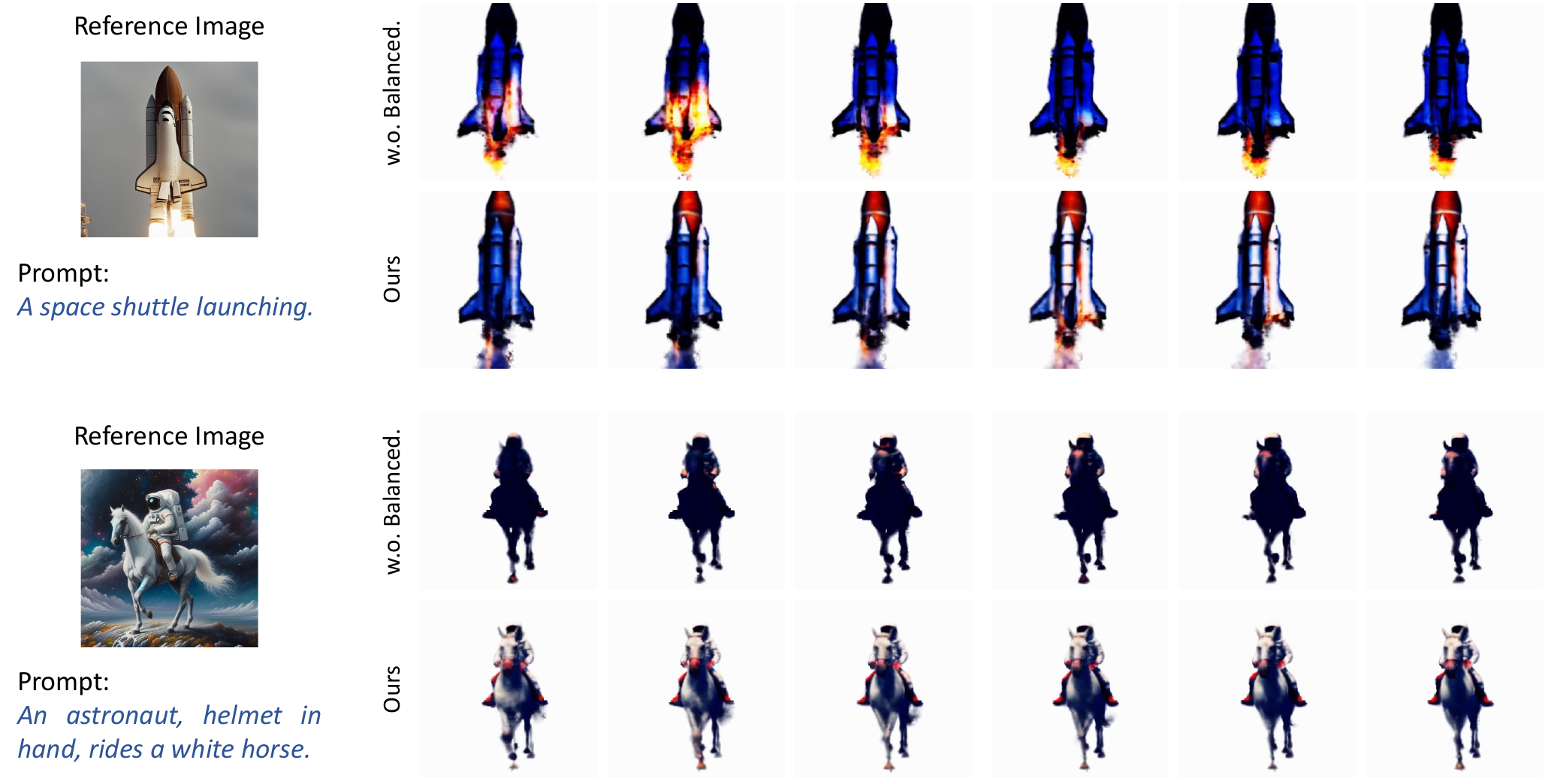}
    \caption{The comparison between the temporal balanced sampling and random sampling. In each example, we display six consecutive frames from the early timesteps. Temporal balanced sampling notably enhances the visual quality of these early frames in the videos.}
    \label{fig:time-sample}
\end{figure*}

\section{Outline}
\begin{itemize}
\item A comparative analysis of our 4D grid encoding and the HexPlane~\cite{hexplane} dynamic NeRF backbone is presented in Sec.~\ref{sec:hexplane};
\item A comparison between our semantic refinement using ControlNet prior~\cite{controlnet} and the refinement based on super-resolution prior is detailed in Sec.~\ref{sec:refine};
\item The qualitative demonstration of the effectiveness of our proposed temporal balanced sampling is provided in Sec.~\ref{sec:sample}.
\end{itemize}

\section{Backbone}
\label{sec:hexplane}
In this paper, we leverage dynamic NeRF with 4D grid encoding to represent the spatio-temporal scene. Specifically, we divide the time dimension evenly into $T$ grids, and for each time grid, we establish a 3D multi-scale grid $V$. The spatio-temporal features are obtained by interpolating two nearest time grids. In contrast, MAV3D~\cite{mav3d} employs HexPlane~\cite{hexplane} dynamic NeRF for 3D video representation. This method maps the X, Y, Z, and time axes onto six 2D planes, fusing these features to calculate density and color.
Our approach is directly compared with HexPlane in Fig.~\ref{fig:hexplane}. \textbf{Note} that for HexPlane, we set the azimuth angle to $45^\circ$ to prevent the reference camera pose from being perpendicular to one of the planes, which could result in the significant Janus problem.
The fish fin in the first example illustrates how 4D grid encoding typically exhibits more motion than HexPlane. Furthermore, the \textbf{back view} of a 4D scene in the second example demonstrates that, despite careful adjustment of the reference camera pose, HexPlane is more prone to the Janus problem compared to 4D grid encoding. Consequently, we choose 4D grid encoding for dynamic scene representation.

\section{Refinement}
\label{sec:refine}

In this paper, we introduce semantic refinement to mitigate the semantic drift associated with video diffusion models and to enhance the resolution of videos generated in the dynamic fine stage. This refinement is achieved through personalized modeling using the ControlNet~\cite{controlnet} diffusion prior. MAV3D~\cite{mav3d} also employs a coarse-to-fine approach, utilizing a super-resolution diffusion prior in their refinement stage to enhance results.

To assess the effectiveness of our approach in addressing semantic drift, we conduct a comparison between our personalized ControlNet prior and an image super-resolution prior\footnote{https://huggingface.co/stabilityai/stable-diffusion-x4-upscaler} in Fig.\ref{fig:sr}. In each example presented in Fig.\ref{fig:sr}, the first row depicts the 3D video as generated in the coarse stage, while the second and third rows show the outcomes following semantic and super-resolution refinement, respectively. It is evident that while super-resolution can improve resolution (as seen in the fourth frame of the first example), it can also amplify errors (such as the back of the fox in the second example) present in the coarse stage, due to its lack of reference image context. In contrast, our semantic refinement not only enhances video quality but also rectifies semantic inaccuracies from the coarse stage.

\section{Temporal Balanced Sampling}
\label{sec:sample}
To enhance the supervision of the reference image and improve the learning of the initial and final time grids, we introduce temporal balanced sampling. 
In Fig.~\ref{fig:time-sample}, we compare this technique with random sampling on two examples of the early timesteps. Since temporal balanced sampling can gather more information from the first frame, our method can present a more accurate appearance in the early stages. 

{
    \small
    \bibliographystyle{ieeenat_fullname}
    \bibliography{reference}
}

\end{document}